\documentclass[letterpaper, 10 pt, conference]{ieeeconf}  

\IEEEoverridecommandlockouts                              
\usepackage{amsmath,amssymb,amsfonts}
\usepackage{booktabs,array,tabularx}
\usepackage{cite}
\usepackage{subcaption}
\usepackage{algorithmic}
\usepackage{graphicx}
\usepackage{textcomp}
\usepackage{xcolor}

\title{\LARGE \bf
Stand, Walk, Navigate: Recovery-Aware Visual Navigation on a Low-Cost Wheeled Quadruped
}

\author{Jans Solano$^{1}$ and Diego Quiroz$^{1}$%
\thanks{$^{1}$The authors are with the Department of Mechatronics Engineering, Pontificia Universidad Católica del Perú. e-mail: {\tt\small \{jans.solano, dequiroz\}@pucp.edu.pe}.}%
\thanks{This paper is part of the IROS full-day workshop ``The Future of Hybrid Mobility: Innovations in Wheeled-Legged Robots'' held on the 24th of October 2025 in Hangzhou, China.}
}

\begin{document}

\maketitle
\thispagestyle{empty}
\pagestyle{empty}

\begin{abstract}
Wheeled–legged robots combine the efficiency of wheels with the obstacle negotiation of legs, yet many state-of-the-art systems rely on costly actuators and sensors, and fall-recovery is seldom integrated, especially for wheeled–legged morphologies. This work presents a recovery-aware visual-inertial navigation system on a low-cost wheeled quadruped. The proposed system leverages vision-based perception from a depth camera and deep reinforcement learning policies for robust locomotion and autonomous recovery from falls across diverse terrains. Simulation experiments show agile mobility with low-torque actuators over irregular terrain and reliably recover from external perturbations and self-induced failures. We further show goal directed navigation in structured indoor spaces with low-cost perception. Overall, this approach lowers the barrier to deploying autonomous navigation and robust locomotion policies in budget-constrained robotic platforms.
\end{abstract}

\section{INTRODUCTION}

Wheeled robots remain the workhorses of mobile robotics due to their high energy efficiency and superior speed on structured, man-made surfaces \cite{siegwart}. Their mechanical design is comparatively simple and reliable, leading to lower manufacturing and maintenance costs, and they enable precise, easily controllable motion suitable for long-duration missions and heavy payloads. However, when the terrain becomes irregular or discontinuous, the advantages of wheels diminish and legged systems take the lead. Legged robots have recently demonstrated remarkable mobility in challenging environments, from climbing staircases \cite{Chamorro2024ReinforcementLF} to exploring underground caves \cite{Olsen2023DesignAE} and forests \cite{Mattamala2024AutonomousFI}. 

However, that versatility comes at a cost: legged robots often operate with higher power consumption (especially off flat terrain), slower average velocities on flat ground when compared to wheels, and require more sophisticated control for gait planning and balance under disturbances \cite{Silva2012ALR}. They also require high-performance actuators and precise sensing to maintain stability and control \cite{7758092}, which increases overall system cost and implementation complexity compared to wheeled platforms. As a result, legged robots are generally more demanding to design, calibrate, and operate reliably over extended use.

Wheeled–legged robots aim to combine the speed and energy efficiency of wheels with the terrain adaptability of legs \cite{Medeiros2020TrajectoryOF, surveyBjelonic}. By driving on flat surfaces and stepping over obstacles as needed, they can traverse easy terrain quickly while overcoming rough sections by legged locomotion. In fact, such hybrids can attain higher average speeds and a lower specific cost of transport on benign terrain compared to purely legged systems \cite{Bjelonic2018KeepRM}. At the same time, they retain the ability to climb steps and even cross gaps that would be impassable for conventional wheeled robots \cite{Klamt2017AnytimeHD}. For example, the ANYmal wheeled-quadruped demonstrated driving at 4 m/s and an 83\% reduction in cost of transport relative to a walking gait \cite{Bjelonic2018KeepRM}. Despite these advantages, practical deployments often hinge on torque-dense, backdrivable actuators and powerful onboard compute, frequently paired with premium perception sensors, thereby driving up overall system cost.
\begin{figure}[t]
    \centering

    \begin{subfigure}{0.48\textwidth}
        \centering
        \includegraphics[width=\linewidth, height=50pt]{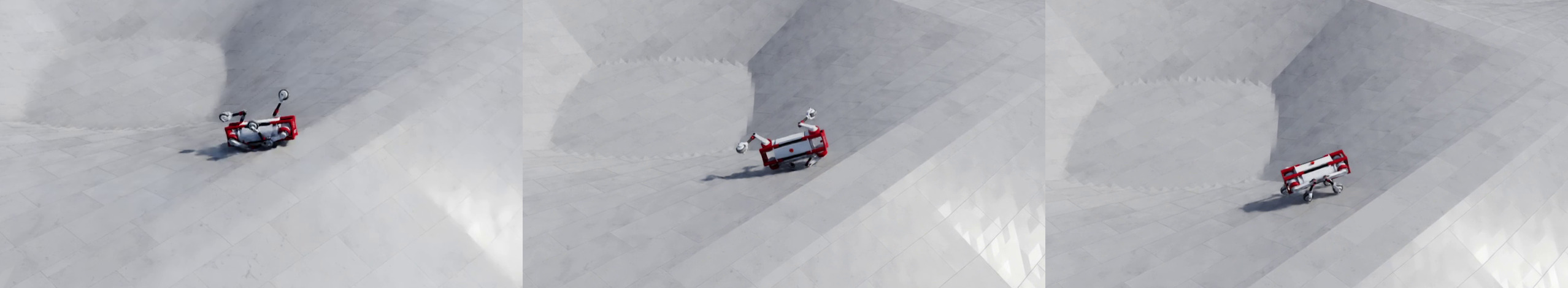}
        \caption{Fall Recovery}
        \label{fig:res_a}
    \end{subfigure}

    \begin{subfigure}{0.48\textwidth}
        \centering
        \includegraphics[width=\linewidth, height=50pt]{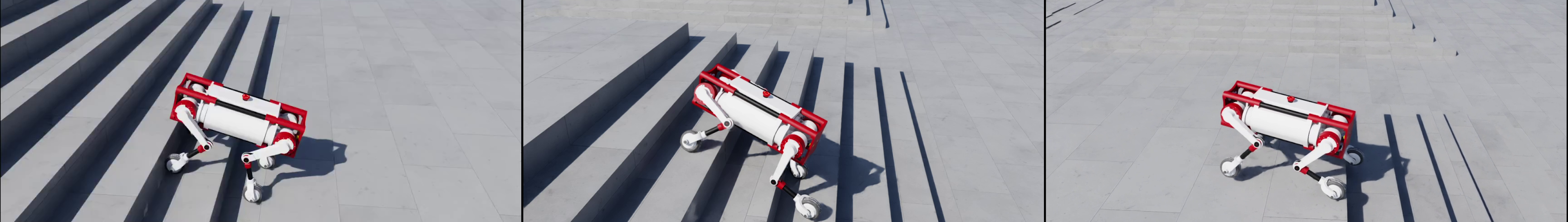}
        \caption{Locomotion}
        \label{fig:res_b}
    \end{subfigure}

    \begin{subfigure}{0.48\textwidth}
        \centering
        \includegraphics[width=\linewidth, height=50pt]{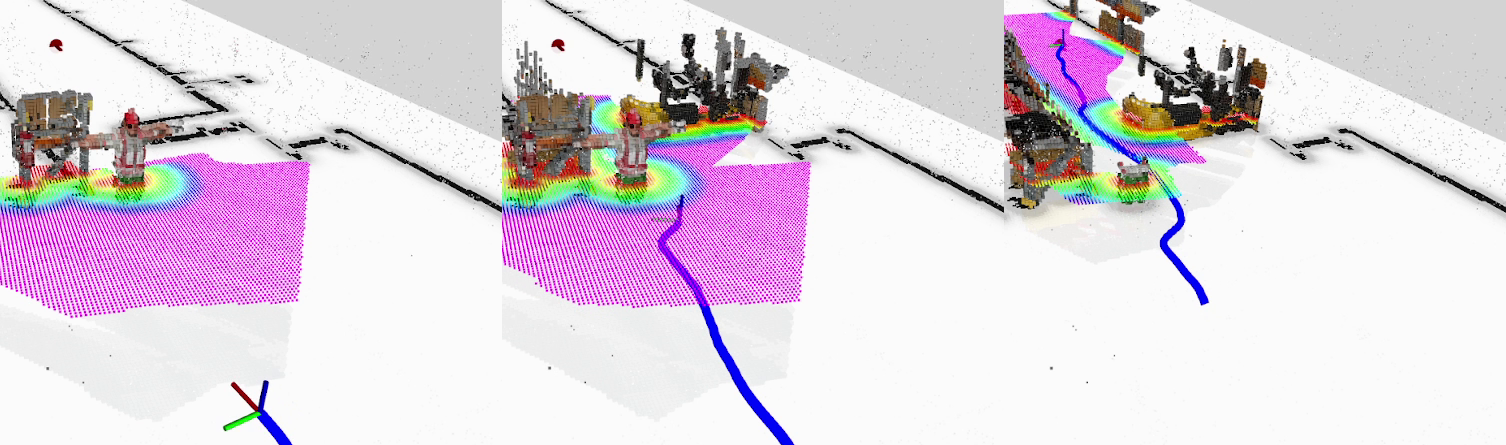}
        \caption{Visual Navigation}
        \label{fig:res_c}
    \end{subfigure}

    \caption{Integrated pipeline: fall recovery, locomotion, and indoor visual navigation.}
    \label{fig:three_stacked}
\end{figure}

Building upon these mechanical advances, recent progress in deep reinforcement learning (DRL) has significantly enhanced the agility and versatility of legged robots. DRL-based controllers have enabled quadrupeds to learn complex locomotion skills directly from experience and to traverse irregular terrain with remarkable robustness \cite{Rudin2021LearningTW}. However, even the most capable locomotion policies cannot fully prevent falls or instability events, making recovery behaviors indispensable for resilient operation. Several recovery strategies have been proposed, including robust posture restoration from arbitrary fallen configurations \cite{Lee2019RobustRC}, state machines utilizing stability feature space \cite{Castano2019DesignAF}, and legged recovery on non-flat terrains through DRL policies \cite{jungyein_thesis}. Yet, most of these efforts focus on purely legged robots or assume relatively flat environments, leaving recovery-aware locomotion for wheeled–legged systems operating on rough terrain underexplored.

These limitations highlight a broader challenge: achieving reliable autonomy in low-cost legged robots. In particular, coupling robust locomotion with perception and navigation remains difficult when hardware constraints limit sensing and computation. While high-end robots rely on precise LiDAR and powerful onboard processing for mapping and localization \cite{Lee2024LearningRA}, affordable platforms often operate with lightweight sensors that degrade state estimation quality. Compounding this, the fast and reactive gaits produced by learned locomotion can destabilize conventional SLAM pipelines: rapid body motions induce sensor jitter, pose drift, and frequent tracking  \cite{aditya2025robustlocalizationmappingnavigation}. These interrelated challenges motivate approaches that explicitly balance agile, learning-based locomotion with resilient perception and low-cost state estimation for autonomous operation.

In this paper, we present an integrated vision-based navigation and locomotion system for a custom-designed, low-cost wheeled quadruped robot evaluated in simulation. The primary contribution of our work is the design and implementation of a complete autonomy stack that enables a wheeled–legged robot to perceive its environment, plan feasible paths, and execute dynamic locomotion maneuvers reliably. Our approach builds upon a state-of-the-art deep reinforcement learning framework for locomotion \cite{fan_ziqi2024robot_lab}, combined with an on-board visual–inertial perception and mapping module. We also develop a recovery algorithm that allows the robot to right itself and continue navigating after potential falls or disturbances.

This work presents a unified pipeline for vision-based navigation in wheeled–legged robots, leveraging modern learning-based locomotion while addressing practical reliability challenges. By integrating these elements into a cohesive framework, we aim to advance the understanding of affordable, fully autonomous legged robot design before real-world deployment. The results provide insights into how navigation and dynamic locomotion can be achieved concurrently on resource-constrained platforms.

\section{METHOD}
We briefly describe the methods underpinning the proposed pipeline, with emphasis on the fall-recovery and navigation modules.
\subsection{Fall Recovery}
We train a fall-recovery policy that brings the robot from fallen or near-fallen states to a stable standing posture. Each episode is initialized with randomized base orientation, base position, joint configuration, and terrain type. To diversify failure modes, the robot was dropped from a random height between 1\,m and 2\,m and then collected a set of 100, 000 poses that are the initial reset states from the robot in each episode.

The policy receives only proprioceptive inputs (Table~\ref{tab:obs-rew}, top): base linear/angular velocity, projected gravity in the base frame \( \mathbf{g}_b \), joint positions/velocities, and the previous action. During training, we also expose a \emph{privileged} signal \( \tilde{h}_{\rm terr}^{\dagger}\!\in\!\mathbb{R}^{187} \) that encodes local terrain-height samples around the robot in a 1.6x1m grid, with 0.1m of resolution. This input is used exclusively by the critic to accelerate learning on rough terrain and is withheld at test time (asymmetric training).
\begin{table}[t]
  \centering
  \caption{Observations (top) and reward terms (bottom)}
  \label{tab:obs-rew}
  \footnotesize
  \renewcommand{\arraystretch}{1.05}

  \begin{tabularx}{\linewidth}{@{}>{\centering\arraybackslash}c
                                   >{\centering\arraybackslash}c
                                   >{\centering\arraybackslash}c
                                   >{\centering\arraybackslash}c@{}}
    \multicolumn{4}{@{}c@{}}{\textit{Observations}} \\
    \toprule
    \textbf{Symbol} & \textbf{Observation} & \textbf{Size} & \textbf{Noise (U[$n_{\min},n_{\max}$])} \\
    \midrule
    $\mathbf{v}_{\rm lin}$ & Base linear velocity                 & $3$              & $[-0.1,\ 0.1]$ \\
    $\boldsymbol{\omega}$  & Base angular velocity                & $3$              & $[-0.2,\ 0.2]$ \\
    $\mathbf{g}_{b}$       & Projected gravity (base frame)       & $3$              & $[-0.05,\ 0.05]$ \\
    $\mathbf{q}$           & Joint positions           & 16 & $[-0.01,\ 0.01]$ \\
    $\dot{\mathbf{q}}$     & Joint velocities          & 16 & $[-1.5,\ 1.5]$ \\
    $\mathbf{a}_{t-1}$     & Last action                          & 16 & (none) \\
    $\tilde{h}_{\rm terr}^{\dagger}$ & Terrain height & 187  & $[-0.1,\ 0.1]$ \\
    \bottomrule
  \end{tabularx}
  {\footnotesize $^{\dagger}$Privileged information.}

  \vspace{4pt}

  \begin{tabularx}{\linewidth}{@{}>{\centering\arraybackslash}c
                                   >{\centering\arraybackslash}c
                                   >{\centering\arraybackslash}X@{}}
    \multicolumn{3}{@{}c@{}}{\textit{Rewards}} \\
    \toprule
    \textbf{Term} & \textbf{Weight} & \textbf{Formula} \\
    \midrule
    $\mathrm{r}_{\text{dof-torq}}$      & $-1{\times}10^{-5}$  & $\lVert \boldsymbol{\tau}\rVert_2^{2}$ \\
    $\mathrm{r}_{\text{dof-acc}}$       & $-2.5{\times}10^{-7}$ & $\lVert \ddot{\mathbf{q}}\rVert_2^{2}$ \\
    $\mathrm{r}_{\text{action-rate}}$   & $-0.1$               & $\lVert \mathbf{a}_{t}-\mathbf{a}_{t-1}\rVert_2^{2}$ \\
    $\mathrm{r}_{\text{wheel-vel}}$     & $-0.01$              & $\displaystyle\sum_{i\in\mathcal{W}} \lvert \dot{q}_i \rvert$ \\
    $\mathrm{r}_{\text{orientation}}$   & $+0.5$               & $\exp(-g_{\text{b,z}}-1),\ \ g_\text{b-z}\in[-1,1]$ \\
    $\mathrm{r}_{\text{joint-track}}$   & $+1.0$               & $\mathbb{1}\{g_{\text{b,z}}<-\tfrac{3}{4}\}\,\exp\!\big(-\lVert \mathbf{q}-\mathbf{q}_{\text{def}}\rVert_{1}\big)$ \\
    $\mathrm{r}_{\text{standing}}$      & $+50.0$              & 
    $\mathbb{1}\{|z + g_\text{b-z}\,h^\star - \overline{z}_{\text{feet}}|<\varepsilon\ \wedge\ \mathrm{contacts}=4  \wedge\ g_{\text{b,z}}<-1+\varepsilon\}
    $\\
    
    \bottomrule
  \end{tabularx}
\end{table}
We detect successful recovery using a contact-aware clearance metric that remains valid on sloped or uneven terrain. Let \( z \) be the base world \( z \) position, \( g_{b,z} \) the base-frame gravity \( z \) component, \( h^\star \) the target base height, and \( \overline{z}_{\text{feet}} \) the mean world \( z \) of feet currently in contact. Define
\begin{equation}
c \;=\; z \;+\; g_{b,z}\,h^\star \;-\; \overline{z}_{\text{feet}}\, .
\end{equation}
We declare success when \(c\) is finite, \(|c|<\varepsilon\), exactly four feet are in contact, and the robot’s orientation is stable (e.g., \(g_{b,z}\le -1+\varepsilon/2\)) with \(\varepsilon=0.1\).

The policy outputs leg joint-position targets and wheel angular-velocity targets. We use a low-torque regime, legs capped at \( \tau_{\max}=25\,\mathrm{Nm} \) and wheels at \( \tau_{\max}=6\,\mathrm{Nm} \) to match the capabilities of low-cost, off-the-shelf actuators. Leg actions are scaled by 0.5 around nominal offsets, while wheel-velocity actions are unit-scaled to keep control stable, limit wheel spin, and remain compatible with budget hardware.
\begin{figure*}[!t]
  \centering
  \includegraphics[width=0.85\textwidth]{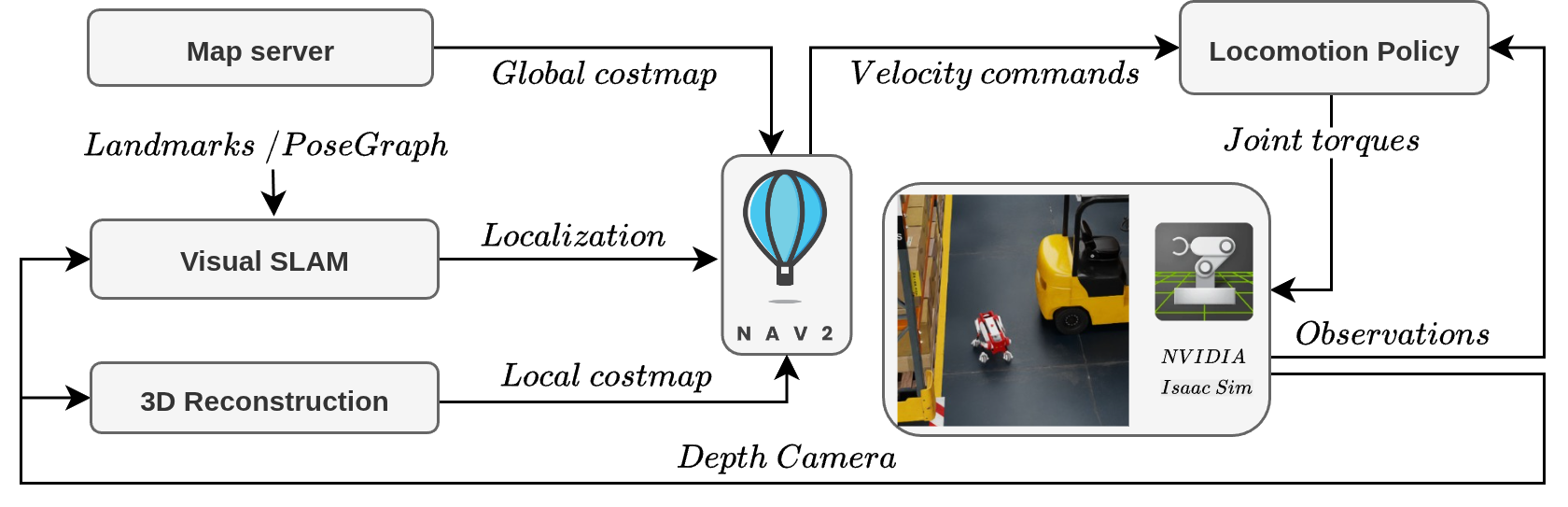}
  \caption{Integrated pipeline for vision-based navigation. NAV2 consumes a global 2D costmap and a local 3D costmap to generate velocity commands that are tracked by the locomotion policy.}
  \label{fig:navigation_overview}
\end{figure*}
We use energy and smoothness regularizers (DOF torque/acceleration and action-rate), a wheel-velocity penalty to discourage inadvertent skidding during the rise, an orientation shaping term based on $g_{b,z}$, and a joint-tracking prior toward a comfortable default pose in highly pitched configurations. A large sparse bonus $\mathrm{r}_{\text{standing}}$ is granted upon meeting the clearance\,+\,contacts condition. Episodes terminate on success or timeout.

The fall-recovery policy was trained using PPO with three hidden layers of sizes [512, 256, 128] and ELU activations. Empirical normalization was disabled to preserve absolute torque magnitudes. Training used an adaptive learning rate of $1\times10^{-3}$, $\gamma=0.99$, $\lambda=0.95$, a clipping parameter of $0.2$, and entropy coefficient of $0.005$. Each update ran for 5 epochs over 4 mini-batches.

\subsection{Locomotion and Navigation}

As illustrated in Fig.~\ref{fig:navigation_overview}, the proposed system couples a learned locomotion controller with a vision-based navigation stack. We train the policy with the framework of \cite{fan_ziqi2024robot_lab}, targeting a maximum commanded speed of $2.0\,\mathrm{m/s}$. A curriculum increases terrain difficulty from flat ground to rough terrain, yielding a policy that remains stable across the operating envelope.

A prior map is built with \emph{nvblox} \cite{Millane2023nvbloxGI} and visual SLAM \cite{Korovko2025cuVSLAMCA}, and is used for localization only mode. The derived 2D occupancy grid provides the global costmap for \emph{NAV2} \cite{Macenski2020TheM2}, while the 3D voxel map supplies the local costmap for obstacle avoidance. The MPPI local planner \cite{WilliamsMPPI} outputs linear and angular velocity commands that the locomotion policy tracks, producing joint torques. In structured indoor environments the robot os commanded with velocities up to $0.7\,\mathrm{m/s}$, safely avoiding obstacles using only depth and IMU sensing. In preliminary tests performed to validate IMU fusion under legged body shake, localization RMSE improved by $\approx 25\%$ overall compared to vision-only.

\section{EXPERIMENTS AND RESULTS}
The reinforcement learning models were trained in Isaac Lab \cite{mittal2023orbit} using an NVIDIA RTX 3060 GPU and subsequently deployed in Isaac Sim \cite{NVIDIA_Isaac_Sim} for evaluation. All experiments were performed in simulation using our low-cost wheeled–legged platform model to assess locomotion and navigation performance.
\begin{table}[t]
\centering
\caption{Recovery performance per terrain type}
\begin{tabular}{lcc}
\toprule
\textbf{Terrain Type} & \textbf{Success Rate (\%)} & \textbf{Avg. Time (s)} \\
\midrule
Stairs   & 90.5 & 0.86 \\
Slopes   & 93.3 & 0.87 \\
Discrete & 86.1 & 0.97 \\
Flat     & 93.4 & 0.90 \\
\bottomrule
\end{tabular}
\label{tab:recovery_performance}
\end{table}

\subsection{Recovery Evaluation}
For evaluation, the robot was exposed to 100,000 trials across various terrain types: stairs with a maximum step height of 0.4 m, slopes up to 45°, and discrete terrains with step variations up to 0.2 m.  
The results are summarized in Table \ref{tab:recovery_performance}. The policy achieved a recovery success rate of up to 93.4 \% on flat terrain and an average of 90.0 \% across irregular terrains. The lowest performance was observed on discrete terrains, likely due to terrain randomness and the absence of exteroceptive sensing during inference, as the policy relied solely on proprioception. These conditions also resulted in a longer average recovery time of 0.97 s.

\subsection{Navigation Performance}
\begin{figure}[t]
\centering
\begin{subfigure}[t]{0.44\linewidth}
  \centering
  \includegraphics[
    width=\linewidth,
    trim=0 0 0 0.72cm, 
    clip
  ]{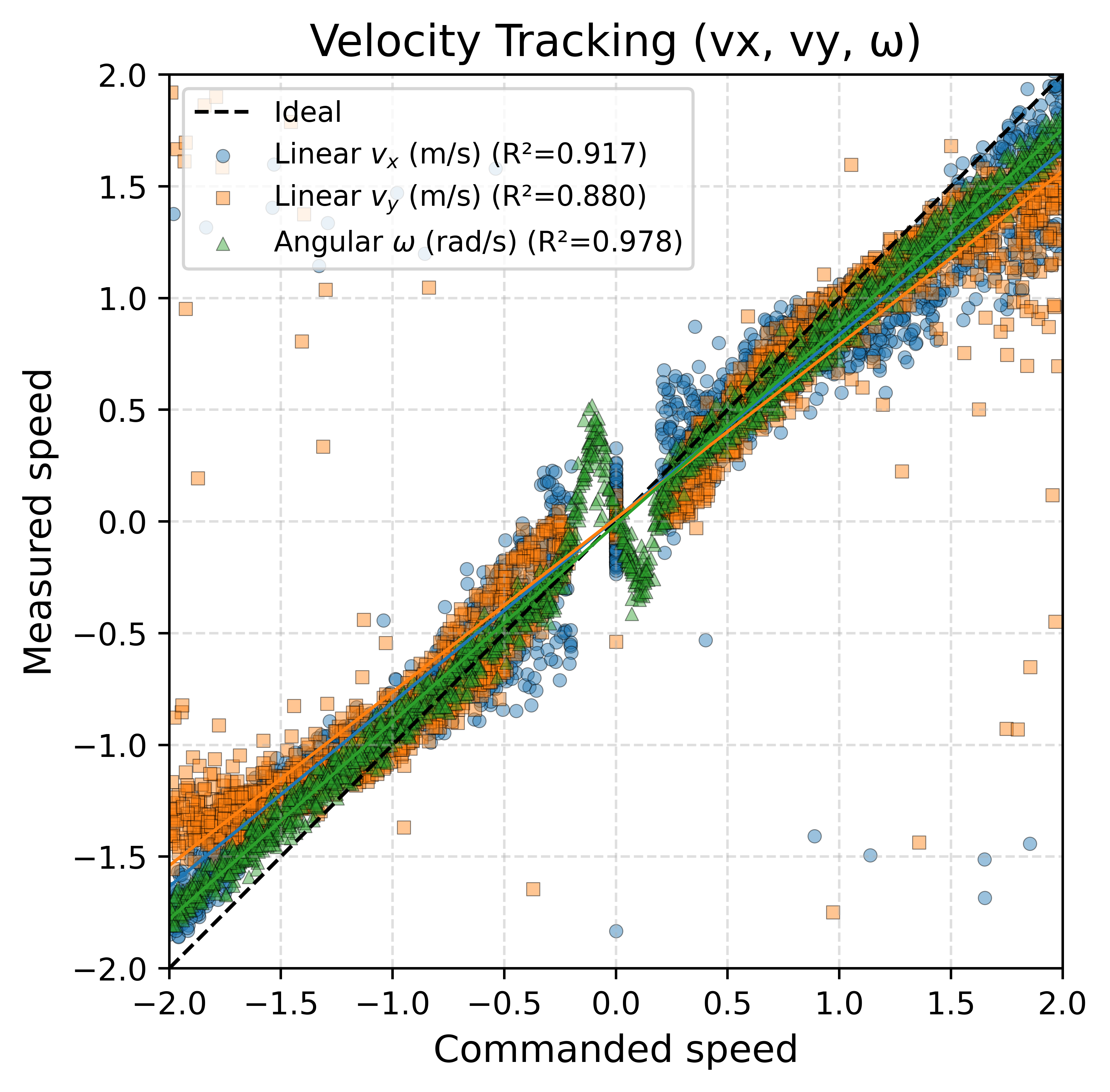}
  \caption{Velocity Tracking.}
  \label{fig:vel_tracking}
\end{subfigure}\hfill
\begin{subfigure}[t]{0.55\linewidth}
  \centering
  \includegraphics[width=\linewidth]{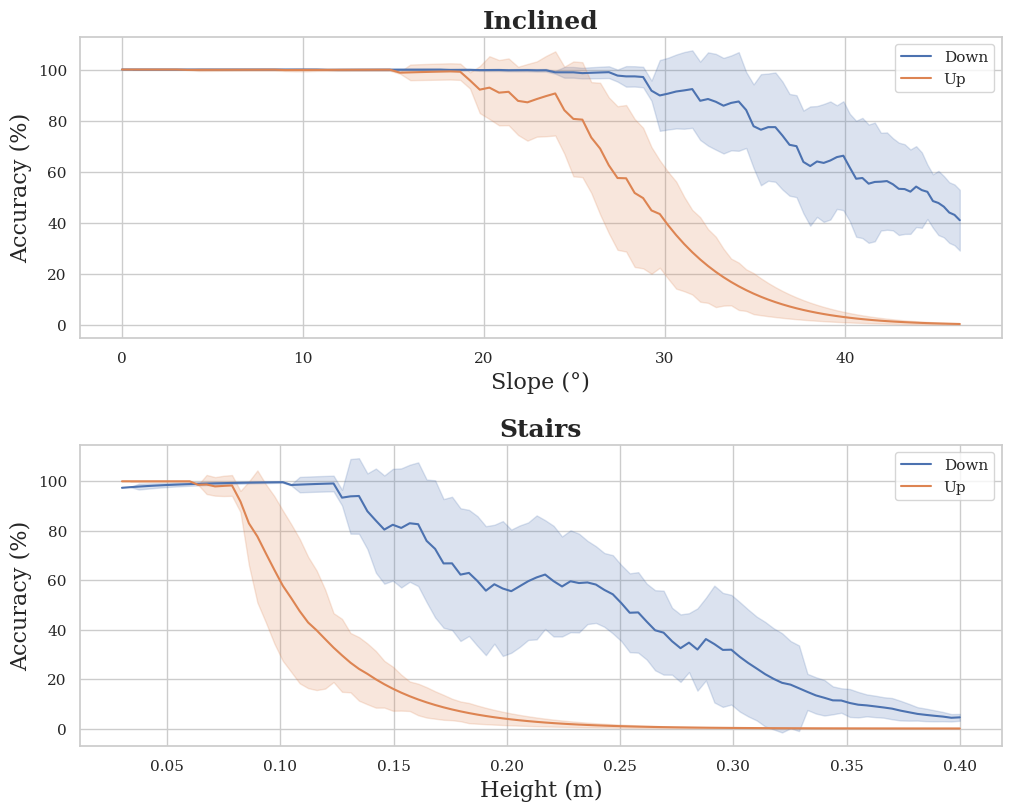}
  \caption{Terrain Accuracy.}
  \label{fig:terrain_acc}
\end{subfigure}
\caption{Locomotion performance results: (a) Velocity Tracking and (b) Terrain Accuracy.}
\label{fig:vel_and_terrain}
\end{figure}
Figure~\ref{fig:vel_and_terrain} summarizes flat terrain speed tracking quality and terrain robustness. For velocity–tracking evaluation, a flat terrain was used where, in each episode, the robot was commanded with random velocities in different directions until it reached the simulation boundaries (\(8 \times 8~\mathrm{m}\)). For rough–terrain locomotion evaluation, the robot started at the center of the arena and was given a maximum forward velocity command of \(2~\mathrm{m/s}\). An episode was considered successful if the robot exited the test area without any contact between the main body and the terrain, or before reaching the time-out limit.

In Figure~\ref{fig:vel_and_terrain}a, the commanded and measured velocities closely align with the ideal line across the entire range, with linear fits showing strong agreement for$v_x$, $v_y$, and $\omega$ (e.g., $R^2\!\approx\!0.92$, $0.88$, and $0.98$, respectively), indicating low bias and good gain adaptation. Figure~\ref{fig:vel_and_terrain}b shows recovery/accuracy as a function of terrain difficulty. Performance remains high for small slopes and low steps, then degrades monotonically as difficulty increases. Descents (blue) consistently outperform ascents (orange), with the difference widening on steeper slopes and higher steps, most likely due to the gravitational factor playing in their favor. The shaded bands (standard estimate) widen at the extreme ranges, reflecting greater variability in the most difficult conditions.

\begin{figure}
\centering
\includegraphics[width=0.8\linewidth]{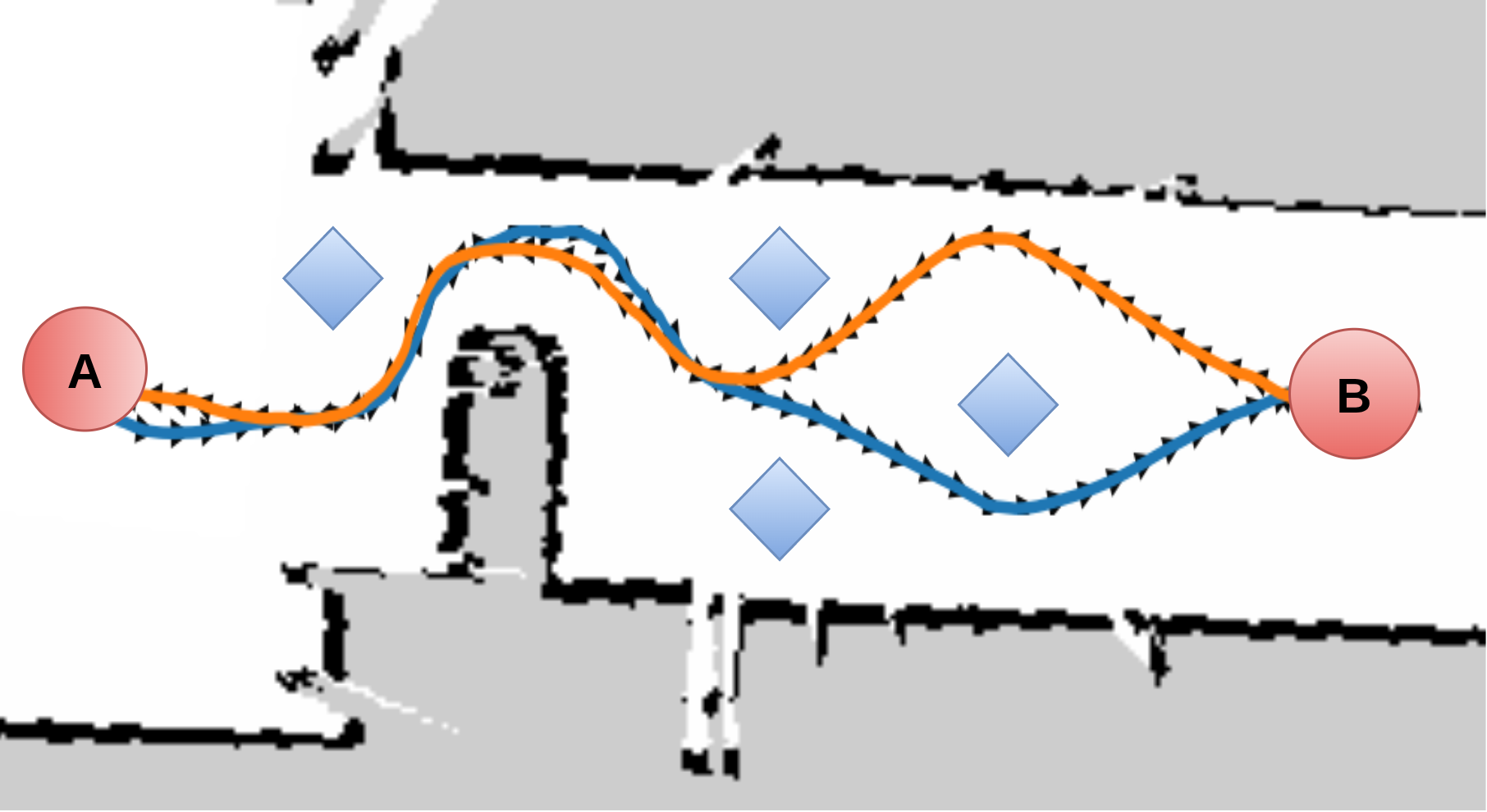}
\caption{Navigation performance methodology.}
\label{fig:nav_perf}
\end{figure}

To evaluate the performance of autonomous navigation, two reference points (A and B) were placed 14\,m apart, as illustrated in Figure~\ref{fig:nav_perf}. The route included three static obstacles (shown as blue rhombuses), and the navigation task was executed over 100 trials, with two representative trajectories displayed. The system achieved a success rate of 98\% demonstrating reliable indoor navigation using only visual perception. The few observed failures mainly resulted from temporary localization losses during traversal.



\section{CONCLUSION AND FUTURE WORK}

We presented a recovery-aware, vision-based navigation system for a low-cost wheeled quadruped robot. The proposed framework integrates DRL-based locomotion and self-righting behaviors with lightweight RGB-D perception and mapping, demonstrating that full autonomy can be achieved without dependence on high-end sensors or actuators. Simulation results highlight robust recovery, agile motion, and reliable goal-directed navigation across diverse indoor environments.

Future work will focus on deploying the learned policies on a physical prototype to validate sim-to-real transfer and extending the navigation framework for robust operation in rough, unstructured outdoor terrains.

\bibliographystyle{IEEEtran} 
\bibliography{IEEEexample.bib}          

\begin{thebibliography}{10}
\providecommand{\url}[1]{#1}
\csname url@samestyle\endcsname
\providecommand{\newblock}{\relax}
\providecommand{\bibinfo}[2]{#2}
\providecommand{\BIBentrySTDinterwordspacing}{\spaceskip=0pt\relax}
\providecommand{\BIBentryALTinterwordstretchfactor}{4}
\providecommand{\BIBentryALTinterwordspacing}{\spaceskip=\fontdimen2\font plus
\BIBentryALTinterwordstretchfactor\fontdimen3\font minus \fontdimen4\font\relax}
\providecommand{\BIBforeignlanguage}[2]{{%
\expandafter\ifx\csname l@#1\endcsname\relax
\typeout{** WARNING: IEEEtran.bst: No hyphenation pattern has been}%
\typeout{** loaded for the language `#1'. Using the pattern for}%
\typeout{** the default language instead.}%
\else
\language=\csname l@#1\endcsname
\fi
#2}}
\providecommand{\BIBdecl}{\relax}
\BIBdecl

\bibitem{siegwart}
R.~Siegwart, I.~R. Nourbakhsh, and D.~Scaramuzza, \emph{Introduction to Autonomous Mobile Robots}, 2nd~ed.\hskip 1em plus 0.5em minus 0.4em\relax The MIT Press, 2011.

\bibitem{Chamorro2024ReinforcementLF}
S.~Chamorro, V.~Klemm, M.~de~la Iglesia~Valls, C.~Pal, and R.~Siegwart, ``Reinforcement learning for blind stair climbing with legged and wheeled-legged robots,'' \emph{2024 IEEE International Conference on Robotics and Automation (ICRA)}, pp. 8081--8087, 2024.

\bibitem{Olsen2023DesignAE}
J.~A. Olsen and K.~Alexis, ``Design and experimental verification of a jumping legged robot for martian lava tube exploration,'' \emph{2023 21st International Conference on Advanced Robotics (ICAR)}, pp. 452--459, 2023.

\bibitem{Mattamala2024AutonomousFI}
M.~Mattamala, N.~Chebrolu, B.~Casseau, L.~Frei{\ss}muth, J.~Frey, T.~Tuna, M.~Hutter, and M.~F. Fallon, ``Autonomous forest inventory with legged robots: System design and field deployment,'' \emph{ArXiv}, vol. abs/2404.14157, 2024.

\bibitem{Silva2012ALR}
M.~F. Silva and J.~A.~T. Machado, ``A literature review on the optimization of legged robots,'' \emph{Journal of Vibration and Control}, vol.~18, pp. 1753 -- 1767, 2012.

\bibitem{7758092}
M.~Hutter, C.~Gehring, D.~Jud, A.~Lauber, C.~D. Bellicoso, V.~Tsounis, J.~Hwangbo, K.~Bodie, P.~Fankhauser, M.~Bloesch, R.~Diethelm, S.~Bachmann, A.~Melzer, and M.~Hoepflinger, ``Anymal - a highly mobile and dynamic quadrupedal robot,'' in \emph{2016 IEEE/RSJ International Conference on Intelligent Robots and Systems (IROS)}, 2016, pp. 38--44.

\bibitem{Medeiros2020TrajectoryOF}
V.~S. Medeiros, E.~Jelavic, M.~Bjelonic, R.~Y. Siegwart, M.~A. Meggiolaro, and M.~Hutter, ``Trajectory optimization for wheeled-legged quadrupedal robots driving in challenging terrain,'' \emph{IEEE Robotics and Automation Letters}, vol.~5, pp. 4172--4179, 2020.

\bibitem{surveyBjelonic}
M.~Bjelonic, V.~Klemm, J.~Lee, and M.~Hutter, ``A survey of wheeled-legged robots,'' in \emph{Robotics in Natural Settings}, J.~M. Cascalho, M.~O. Tokhi, M.~F. Silva, A.~Mendes, K.~Goher, and M.~Funk, Eds.\hskip 1em plus 0.5em minus 0.4em\relax Cham: Springer International Publishing, 2023, pp. 83--94.

\bibitem{Bjelonic2018KeepRM}
M.~Bjelonic, C.~D. Bellicoso, Y.~de~Viragh, D.~V. Sako, F.~Tresoldi, F.~Jenelten, M.~Hutter, and N.~G. Tsagarakis, ``Keep rollin’—whole-body motion control and planning for wheeled quadrupedal robots,'' \emph{IEEE Robotics and Automation Letters}, vol.~4, pp. 2116--2123, 2018.

\bibitem{Klamt2017AnytimeHD}
T.~Klamt and S.~Behnke, ``Anytime hybrid driving-stepping locomotion planning,'' \emph{2017 IEEE/RSJ International Conference on Intelligent Robots and Systems (IROS)}, pp. 4444--4451, 2017.

\bibitem{Rudin2021LearningTW}
N.~Rudin, D.~Hoeller, P.~Reist, and M.~Hutter, ``Learning to walk in minutes using massively parallel deep reinforcement learning,'' \emph{ArXiv}, vol. abs/2109.11978, 2021.

\bibitem{Lee2019RobustRC}
J.~Lee, J.~Hwangbo, and M.~Hutter, ``Robust recovery controller for a quadrupedal robot using deep reinforcement learning,'' \emph{ArXiv}, vol. abs/1901.07517, 2019.

\bibitem{Castano2019DesignAF}
J.~A. Castano, C.~Zhou, and N.~G. Tsagarakis, ``Design a fall recovery strategy for a wheel-legged quadruped robot using stability feature space,'' \emph{2019 IEEE International Conference on Robotics and Biomimetics (ROBIO)}, pp. 41--46, 2019.

\bibitem{jungyein_thesis}
J.~Lee, ``Deep reinforcement learning for fall-recovery control on non-flat terrain of quadruped robots,'' Master's thesis, 2024, Sungkyunkwan University.

\bibitem{Lee2024LearningRA}
J.~Lee, M.~Bjelonic, A.~Reske, L.~Wellhausen, T.~Miki, and M.~Hutter, ``Learning robust autonomous navigation and locomotion for wheeled-legged robots,'' \emph{Science Robotics}, vol.~9, 2024.

\bibitem{aditya2025robustlocalizationmappingnavigation}
\BIBentryALTinterwordspacing
D.~Aditya, J.~Huang, N.~Bohlinger, P.~Kicki, K.~Walas, J.~Peters, M.~Luperto, and D.~Tateo, ``Robust localization, mapping, and navigation for quadruped robots,'' 2025. [Online]. Available: \url{https://arxiv.org/abs/2505.02272}
\BIBentrySTDinterwordspacing

\bibitem{fan_ziqi2024robot_lab}
\BIBentryALTinterwordspacing
Z.~Fan, ``robot\_lab: Rl extension library for robots, based on isaaclab.'' 2024. [Online]. Available: \url{https://github.com/fan-ziqi/robot\_lab}
\BIBentrySTDinterwordspacing

\bibitem{Millane2023nvbloxGI}
A.~Millane, H.~Oleynikova, E.~Wirbel, R.~Steiner, V.~Ramasamy, D.~Tingdahl, and R.~Siegwart, ``nvblox: Gpu-accelerated incremental signed distance field mapping,'' \emph{2024 IEEE International Conference on Robotics and Automation (ICRA)}, pp. 2698--2705, 2023.

\bibitem{Korovko2025cuVSLAMCA}
A.~Korovko, D.~Slepichev, A.~Efitorov, A.~Dzhumamuratova, V.~Kuznetsov, H.~Rabeti, J.~Biswas, and S.~Pouya, ``cuvslam: Cuda accelerated visual odometry and mapping,'' \emph{ArXiv}, vol. abs/2506.04359, 2025.

\bibitem{Macenski2020TheM2}
S.~Macenski, F.~J.~P. Mart'in, R.~White, and J.~G. Clavero, ``The marathon 2: A navigation system,'' \emph{2020 IEEE/RSJ International Conference on Intelligent Robots and Systems (IROS)}, pp. 2718--2725, 2020.

\bibitem{WilliamsMPPI}
G.~Williams, P.~Drews, B.~Goldfain, J.~M. Rehg, and E.~A. Theodorou, ``Aggressive driving with model predictive path integral control,'' in \emph{2016 IEEE International Conference on Robotics and Automation (ICRA)}, 2016, pp. 1433--1440.

\bibitem{mittal2023orbit}
M.~Mittal, C.~Yu, Q.~Yu, J.~Liu, N.~Rudin, D.~Hoeller, J.~L. Yuan, R.~Singh, Y.~Guo, H.~Mazhar, A.~Mandlekar, B.~Babich, G.~State, M.~Hutter, and A.~Garg, ``Orbit: A unified simulation framework for interactive robot learning environments,'' \emph{IEEE Robotics and Automation Letters}, vol.~8, no.~6, pp. 3740--3747, 2023.

\bibitem{NVIDIA_Isaac_Sim}
\BIBentryALTinterwordspacing
{NVIDIA}, ``{Isaac Sim}.'' [Online]. Available: \url{https://github.com/isaac-sim/IsaacSim}
\BIBentrySTDinterwordspacing

\end{thebibliography}

\end{document}